\documentclass[a4paper,conference]{IEEEtran}
\usepackage{tikz} 
\usepackage{pgfplots} 
\usepackage{cuted}
\usepackage{capt-of}
\usepackage{graphicx}
\usepackage{epstopdf}
\usepackage{subfig}
\usepackage{amssymb}
\usepackage{color}
\usepackage{epsfig} 
\usepackage{amsmath}
\usepackage{balance}
\usepackage{url}
\usepackage{ctable}
\usepackage{floatrow}
\usepackage{tabularx}
\usepackage{multirow}
\usepackage{array}

\usepackage{comment}

\definecolor{myRed}{rgb}{0.74,0.53,0.61}
\definecolor{myOrange}{rgb}{0.74,0.59,0.48}
\definecolor{myYellow}{rgb}{0.77,0.73,0.40}
\definecolor{myGreen}{rgb}{0.52,0.75,0.49}
\definecolor{myBlue}{rgb}{0.45,0.6,0.68}
\definecolor{myPurple}{rgb}{0.57,0.53,0.63}

\usepackage{hyperref}
\hypersetup{
    colorlinks=true,
    linkcolor=blue,
    filecolor=magenta,      
    urlcolor=cyan,
}
\usepackage{soul}
\usepackage{xcolor}
\usepackage{graphicx}
\usepackage{amsmath}
\usepackage{pifont}

\newcommand{\ie}{\emph{i.e.},}
\newcommand{\etal}{\emph{et~al.}}
\newcommand{\cmark}{\ding{51}}
\newcommand{\xmark}{\ding{55}}
\begin{document}
%
\title{\LARGE \bf{Temporally Coherent Embeddings for Self-Supervised Video Representation Learning}}

\author{\IEEEauthorblockN{Joshua Knights, Ben Harwood, Daniel Ward, Anthony Vanderkop, Olivia Mackenzie-Ross, Peyman Moghadam} \\
\IEEEauthorblockA{Robotics and Autonomous Systems, Data61 CSIRO,  Brisbane, QLD 4069, Australia\\
{\tt\small \{firstname.lastname\}@csiro.au}}}

\maketitle

\begin{strip}\centering
\includegraphics[width=0.30\textwidth]{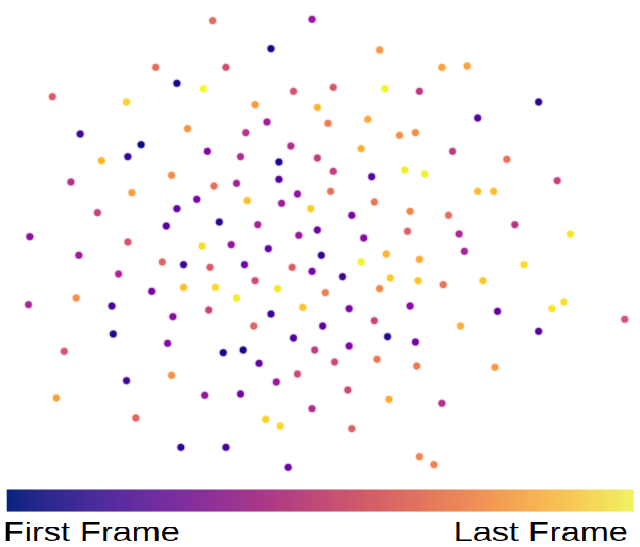}
\includegraphics[width=0.30\textwidth]{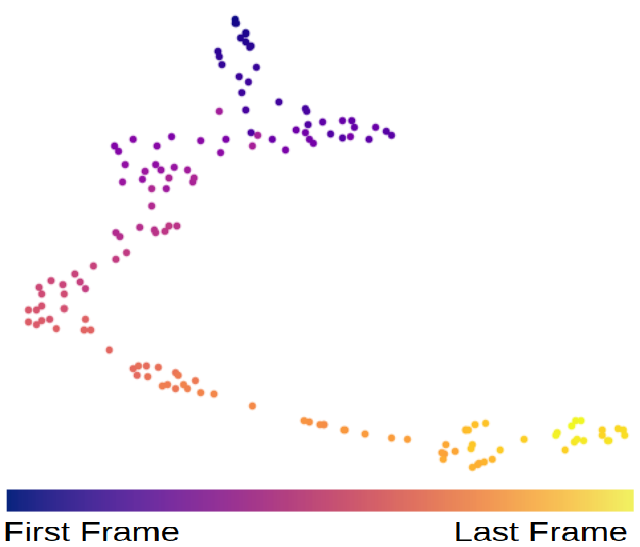}
\includegraphics[width=0.30\textwidth]{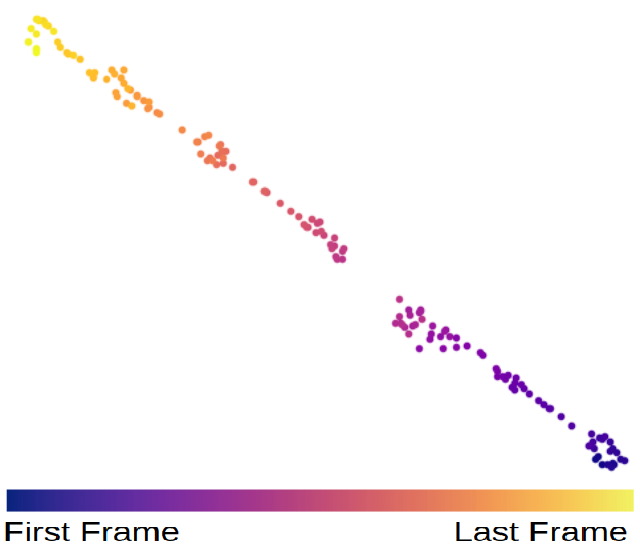} \\
\begin{tabular}{>{\centering}p{0.3\textwidth}>{\centering}p{0.3\textwidth}>{\centering}p{0.3\textwidth}}
(a) Random Initialisation & (b) TCE Epoch 25 & (c) TCE Epoch 50
\end{tabular}
\captionof{figure}{\textit{TCE: Temporally Coherent Embeddings} of frames from a single video visualised using t-SNE, pre-trained on Kinetics400.  
\label{fig:training_prog}}
\end{strip}

\begin{abstract}
\label{sec:abstract}
This paper presents \textit{TCE: Temporally Coherent Embeddings} for self-supervised video representation learning. The proposed method exploits inherent structure of unlabeled video data to explicitly enforce temporal coherency in the embedding space, rather than indirectly learning it through ranking or predictive proxy tasks. In the same way that high-level visual information in the world changes smoothly, we believe that nearby frames in learned representations will benefit from demonstrating similar properties. Using this assumption, we train our \textit{TCE} model to encode videos such that adjacent frames exist close to each other and videos are separated from one another. Using \textit{TCE} we learn robust representations from large quantities of unlabeled video data. We thoroughly analyse and evaluate our self-supervised learned \textit{TCE} models on a downstream task of video action recognition using multiple challenging benchmarks (Kinetics400, UCF101, HMDB51). With a simple but effective 2D-CNN backbone and only RGB stream inputs, \textit{TCE} pre-trained representations outperform all previous self-supervised 2D-CNN and 3D-CNN pre-trained on UCF101. The code and pre-trained models for this paper can be downloaded at: \href{https://github.com/csiro-robotics/tce.git}{https://github.com/csiro-robotics/TCE}

\end{abstract}

\IEEEpeerreviewmaketitle
\section{Introduction}
\label{sec:intro}

\begin{figure*}[t]
\centering  
\includegraphics[width=.7\linewidth]{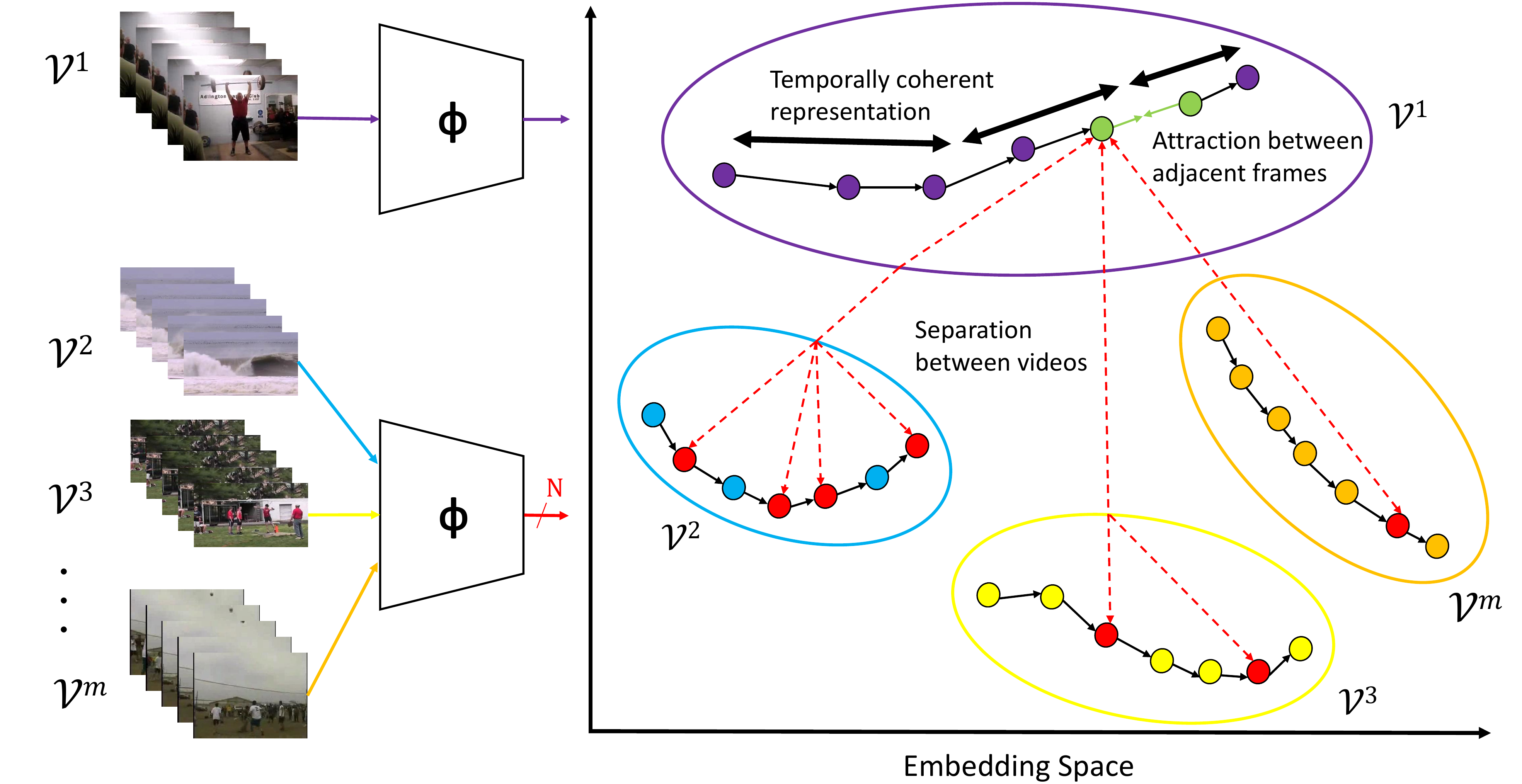}
\caption{Overview of \textit{TCE: Temporally Coherent Embeddings} for self-supervised video representation learning. At each step relative attraction and separation is achieved by computing our temporal coherency objective using an anchor frame, an adjacent frame and \textit{N} negative frames sampled from other videos.} \label{fig:a}
\end{figure*}

Self-Supervised Learning (SSL) is a new and promising paradigm. In SSL, a model is trained on unlabeled data and supervised with a learning signal constructed from inherent structure in the training samples. Robust representations can be learned from enormous amounts of unlabeled data using SSL. These methods are often pre-trained on large unlabeled datasets with specific upstream (\ie{}~\textit{proxy} or \textit{pretext}) tasks and then fine-tuned to adapt to specific downstream tasks.

Video data is an appealing datatype as many image level proxy tasks are applicable and the inherent temporal signal provides an additional source of supervision. Leveraging the temporal aspect of video data, many methods attempt to predict future frames~\cite{lotter2016deep,mathieu2015deep,srivastava2015unsupervised,vondrick2016generating}. However, these methods suffer from wasted model capacity attempting to reconstruct information such as texture or lighting changes which are not transferable to many downstream tasks, \textit{e.g.} action recognition.
Other methods have avoided the need for pixel level reconstruction by predicting latent representations of future frames~\cite{han2019video,vondrick2016anticipating}. Following the belief that spatiotemporal convolutions in 3D-CNNs are vital for video processing~\cite{tran2018closer}, many of the best performing methods for learning from video data rely on 3D-CNN architectures and hence, require large amounts of data and computing resources for training. Ranking methods avoid reconstruction entirely by instead solving proxy tasks aimed at recovering the temporal frame ordering~\cite{el2019skip,fernando2017self,kim2019self,lee2017unsupervised,misra2016shuffle,xu2019self}.

%
%
%

Inspired by recent advances in self-supervised learning in the image domain \cite{bachman2019learning, chen2020simple} we introduce \textit{Temporally Coherent Embeddings (TCE)}, a spatiotemporal approach to self-supervised learning. Our approach is driven by a novel objective function designed to learn a temporally coherent embedding space from unlabeled videos as seen in Figure~\ref{fig:training_prog}. Our \textit{temporal coherency objective} enables a network to learn a temporally coherent embedding space where adjacent frames exist close to each other along a smooth trajectory and distinct videos are embedded with separation between them.


By explicitly optimising for temporal coherency in the embedding space using a 2D-CNN, we avoid the challenges of pixel level reconstruction losses and the computationally expensive 3D-CNN architectures. We argue that equivalent performance can be achieved using 2D-CNNs and an adequately defined loss function. This results in faster training, lower resource requirements and faster inference for our method. We also believe that our temporal coherency objective will better generalise and subsequently translate to downstream tasks when compared to other upstream proxy tasks, such as frame ranking. By encouraging the formation of a structured, intuitive and temporally coherent learned representation we demonstrate strong results on the downstream task of action recognition when compared to other methods.
We summarise our contributions in this paper as follows:

\begin{itemize}
    \item We propose a novel temporal coherency objective for self-supervised video representation learning. This objective exploits the structure of video data to explicitly enforce temporal coherency in the embedding, rather than indirectly learning it through ranking or predictive tasks.
    \item We quantitatively evaluate the effectiveness of our temporal coherency objective at encoding spatiotemporal information both by comparing RGB and stack-of-differences inputs and through the exploration of a higher order objective
    \item We demonstrate that, contrary to the current direction of the field, our 2D-CNNs can learn robust spatiotemporal embeddings for downstream action recognition tasks without relying on data and compute hungry 3D-CNNs.
    \item We formulate a novel approach for semi-hard mining of negative samples during self-supervised learning on large datasets and apply this method when training on Kinetics400.
    \item TCE outperforms all previous state-of-the-art self-supervised action recognition methods that have been pre-trained on UCF101.
    For methods pre-trained on Kinetics400, our approach also outperforms all but one of the competing methods on UCF101 and achieves state-of-the-art for HMDB51.
\end{itemize}

\section{Related Works} \label{sec: Related Works}
%

Our approach relates to contrastive SSL methods which build representations by modelling the differences and similarities between two or more inputs. Here the supervisory signal is obtained by contrastively ranking similar (positive) samples with respect to negative examples. The key differences are that the loss is computed at a feature level and negative examples are required to contrast against. Common image examples include contrasting between different view points~\cite{sermanet2018time}, data modalities~\cite{tian2019contrastive}. and colour spaces~\cite{tian2019contrastive}.

Other work, more closely related to ours, learns from the temporal dynamics of video data. Many approaches attempt to generate future video frames~\cite{lotter2016deep,mathieu2015deep,srivastava2015unsupervised,vondrick2016generating} but suffer from the aforementioned pixel level reconstruction loss limitations~\cite{zhang2017split,han2019video}.  \cite{gan2018geometry} use geometric information such as optical flow and left-right disparity to pre-train the network backbone. Others classified video sequences as having the correct frame ordering~\cite{fernando2017self,misra2016shuffle} or sorting the frames.   Of the sorting methods, those employing 3D-CNNs~\cite{el2019skip,kim2019self,xu2019self, han2019video} require significantly more computational resources but outperform their 2D-CNN counterparts~\cite{lee2017unsupervised}. Many recent works leverage 3D-CNNs which are pre-trained on the much larger Kinetics400 dataset. Contrary to prior 3D-CNN works, \textit{TCE} demonstrates that competitive spatiotemporal features can be learned using 2D-CNNs when optimising for temporal coherency in the embedding space.

Avoiding computationally expensive 3D-CNNs, complex frame sorting and pixel level loss limitations, we explicitly optimise for a representation which enforces cosine similarity between frames by contrastively clustering the embeddings of neighbouring frames. This results in a temporally coherent embedding space (Fig~\ref{fig:a}) where different videos are separated and individual frames are attracted. 
Temporal coherency was explored by \cite{jayaraman2016slow} as an auxiliary signal for learning from videos. However, their formulation only utilises a triplet loss and operates in a semi-supervised setting, whereas we use a combination of contrastive clustering with many negatives and a semi-hard negative mining curriculum to extend our approach to learning in a totally self-supervised manner.
This approach is motivated by recent results which achieve performance improvements proportional to the number of negatives used~\cite{bachman2019learning,tian2019contrastive,tschannen2019mutual}.
Other works have also explored coherency in terms of frame correspondences. \cite{milbich2017unsupervised} solve a pairwise sequence matching problem using triplet loss and bounding box annotations for pose estimation in video. Wang \etal{}~\cite{wang2015unsupervised} uses a triplet loss formulation to learn patchwise frame correspondences. This work, to the best of our knowledge, is the first to utilise temporal coherency as the only supervisory signal.

\section{Methodology}

We propose a simple, yet effective framework to learn a temporally coherent embedding space from unlabeled videos. The mathematical formulation of our method, \textit{TCE: Temporally Coherent Embeddings} for self-supervised video representation learning, is explained in this section.

\subsection{Temporal Coherency Training}
    
The goal of our proposed method, \textit{TCE}, is to explicitly enforce coherency in the embedding space by encouraging similarity in the embeddings of temporally adjacent frames from a video without any labels. 

To develop representations which are coherent in time, we seek to learn an embedding function $f(.)$ which transforms a video frame $x^i_t$ from pixel-space into a lower dimensional embedding space. We adopt the shorthand notation $f(x_t^i) := f_t^i$ for the transformed frame.

We define temporal coherency as minimisation of the temporal derivatives of the representations in embedding space. First-order temporal coherency in the embedding space is thus achieved when $\partial f_t / \partial t \approx 0$. 
To the first order, temporal coherency is achieved by maximising a similarity function $s(f_{t+1}^i,f_t^i)$ between two temporally adjacent frames in the same video, as $\partial f_t / \partial t \propto f_{t+1} - f_t$.  For this purpose, we use the cosine similarity function.
A trivial solution to this optimisation goal is apparent: an embedding function which simply maps all inputs to the same point in the embedding space. To avoid this trivial solution, we define a loss formulation which enforces high cosine similarity between the embeddings of neighbouring frames while enforcing low cosine similarity against a set of \textit{negative} frame embeddings.

Consider a video dataset where each video $\mathcal{V}^i$ contains $T^i$ frames $\left\{ x^i_1, x^i_2, ..., x^i_{T^i} \right\}$. 
We consider a pair of temporally neighbouring frames from one video as positive examples, and consider all frames from other videos to be negative examples. We sample these negative examples to form a set $\mathcal{N}$ containing $N$ frames. 
We adopt a standard cross-entropy loss in Equation \ref{eq:cross-entropy} which is minimised when $s(f_t,f_{t+1})$ is large and $s(f_t, f_n)$ is small for all $x_n \in \mathcal{N}$. 

\begin{align}
    \mathcal{L}_{1^\mathrm{st}}(x^i_{t+1},x^i_{t},\mathcal{N}) = -\mathbb{E} \left[ \log{} \frac{ e^{s(f_t,f_{t+1})} } { e^{s_1(f_t,f_{t+1})}+\sum_{\mathcal{N}}{e^{s(f_t, f_n)}} } \right]
    \label{eq:cross-entropy}
\end{align}

Minimising this loss function is analogous to training a classifier to correctly select the positive example from all negative examples in $\mathcal{N}$. This first order coherency objective will encourage neighbouring video frames to cluster in representation space because it penalises large distances between frames in the embedding space. Additional temporal structure in the embeddings can also be captured through higher order coherency. In Section \ref{sec:CohenenceEval} we explore the use of a second-order temporal coherency objective.

\subsection{Leveraging Multiple Negative Examples with NCE}

For large numbers of negative examples, calculating the normalization factor for the full softmax distribution in Equation \ref{eq:cross-entropy} can prove computationally intractable \cite{mnih2013learning}. Noise Contrastive Estimation (NCE) \cite{gutmann2010noise} is a computationally efficient means of estimating unnormalised statistical models and performing logistic regression to discriminate between observed data and a noise distribution. In this case, discriminating between the positive and negative examples. 
    





The NCE based approximation of the optimization goal of the model can be adapted from Equation \ref{eq:cross-entropy} as
\begin{align}
    \mathcal{L}_{NCE} = -\mathbb{E}_{x, x_p} \left[ \log P(C | x; x_p) \right] \, + & \nonumber \\
    N \,.\, \mathbb{E}_{x_n \in \mathcal{N}} & \left[ \log P(\tilde{C} | x; x_n)  \right]
\end{align}

where $P(\tilde{C} | x; x_n) = 1 - P(C | x; x_n)$ is the probability of correctly classifying a sample from a uniform noise distribution. 
  
\subsection{Annealing as a Semi-Hard Mining Curriculum}
\label{sec:mining}

While increasing the number of negative examples can lead to improved performance on a number of tasks, the vanishing gradient problem is still a concern when training on larger datasets that contain a broad range of visually distinct videos. In these training environments, randomly sampling negative frames will result in a reduced rate of learning once the majority of negatives have been address by the loss function.

Drawing both from annealing methods and semi-hard mining approaches in the supervised domain \cite{schroff2015facenet,harwood2017smart} we formulate a novel approach for negative selection during self-supervised learning on large datasets. Our goal is to define a fine-grained online training curriculum that mines increasingly more challenging negatives in order to continuously produce useful training gradients for self-supervised learning. As such, we define hyperspherical boundaries that are centred on each embedded video frame and have decaying radii given by:

\begin{align}
    r\left(t\right) = r_0 + (r_E - r_0) \left(1 - e^{-\frac{5t}{E}}\right)
    \label{eq:mining}
\end{align}

with $t$ a fractional representation of the current progress through training epochs, $r_0$ the initial radius at epoch $0$ and $r_E$ the radius at final epoch $E$. When selecting negative examples using these boundaries we begin by selecting the closest negatives that are embedded outside of a particular hypersphere. If during the early epochs of training there are fewer mined negatives than the amount required for training, then the remainder are selected with random sampling inside the hypersphere.  This approach also addresses the potential drawback of sampling negatives from highly similar videos in the dataset, as frames from said videos would not be considered until very late in the training process if at all.
\section{Experimental Setup}
\label{sec:experimental}
In this section we describe the specific experimental parameters that we have applied in generating the results we present in Section \ref{sec:ResultsAnalysis} below.

\subsection{Datasets and Augmentation}
\label{subsec:datasets}
In this paper, our pre-training and evaluation is focused on video action recognition datasets UCF101 \cite{soomro2012UCF101}, HMDB51 \cite{kuehne2011hmdb} and Kinetics400 \cite{kay2017kinetics}. UCF101 contains 13K videos split between 101 action classes, and is used for pre-training several of our ablation studies. HMDB51 contains 7K videos split between 51 action classes and Kinetics400 contains 306K videos split between 400 action classes. 
During pre-training we randomly crop a $224\times224$ window from each video frame as an input to a network. We augment these inputs by applying random crop, random horizontal flip, random grey and color jittering.

\subsection{Network Architecture}
We use 2D ResNet-18 \cite{he2016deep} as the backbone for our network architecture unless otherwise stated for our experiments.  The final convolutional layer of the network is flattened and passed through a single fully-connected layer to produce a 128-dimensional feature vector for the network output.

\subsection{Self-Supervised Pre-training}
For our UCF101 experiments, we initialise our networks with random weights and train for a total of $E=9$ epochs on 4 Tesla-V100 GPUs. Here one epoch consists of every frame in UCF101 being used as an anchor example. For our Kinetics400 experiments we instead train for $E=50$ epochs, however for these epochs we are only using a single randomly selected anchor from each pre-training video. All pre-training is performed using a stochastic gradient descent optimiser and a batch size of 100. We set an initial learning rate of $0.03$ and then reduce it by a factor of 10 after 5 epochs for UCF101 and 25 epochs for Kinetics400. In addition to the data augmentation described above, we follow \cite{gidaris2018unsupervised} in pre-training with a rotation auxiliary task to predict a rotation angle from the set $\left[0^{\circ},90^{\circ},180^{\circ},270^{\circ}\right]$ to avoid trivial solutions.

\subsection{Negative Selection}
All of our experiments are run using $\mathcal{N} = 8192$ negative samples for optimising our first order coherency objective. We maintain a memory bank to store the embedded representation of each frame in the dataset for UFC101 and for the most recent anchor in each video for Kinetics400. This allows us to efficiently retrieve noisy samples without re-computing their embeddings. As such, the memory bank is dynamically updated with the new anchor embeddings on every forward pass of the network. Due to our use of cosine similarity, our semi-hard mining limits from Equation \ref{eq:mining} are set to $r_0 = -1$ and $r_E = 1$.

\subsection{Evaluation on Action Recognition}
After self-supervised pre-training, we evaluate our models by their performance on the downstream tasks of action recognition on UCF101 and HMDB51. In general we report top-1 accuracy for the downstream task on the first testing split of the dataset. However, for our comparison with state-of-the-art in Table \ref{tab:soa_table} we report the average accuracy over three splits. We replace the final fully-connected layer of our pre-trained networks with a new fully-connected layer that has a dimensionality equal to the number of classes in the evaluation dataset. The network is then fine-tuned using 4 Tesla-V100 GPUs and stochastic gradient descent for 600 epochs, with a learning rate of 0.05 decayed by a factor of 0.1 at 375 epochs. For networks pre-trained on Kinetics400 we fine-tune for an additional 300 epochs, decaying the learning rate again at 600 epochs, and for our HMDB51 results we use a dropout of 0.9 for the fully connected layer. During fine-tuning, each video is divided into three equal portions and a frame is randomly sampled from each for passing through the network. The average output feature of these three frames is then used as the final prediction. Then during evaluation, nineteen evenly spaced frames are sampled from each video and the softmax averaged output features from these frames is then used to determine the network's prediction. Our reported results are taken from the highest scoring epoch.
    
    



\section{Results and Analysis} 
\label{sec:ResultsAnalysis}

In this section we evaluate and analyse the performance of our proposed method in order to better understand what role temporal consistency plays in both forming embeddings and for downstream action recognition tasks~\footnote{Supplementary materials at \href{https://csiro-robotics.github.io/TCE-Webpage}{https://csiro-robotics.github.io/TCE-Webpage}}.
In Section \ref{sec:Auxiliary} we demonstrate the importance of using an auxiliary task to avoid learning trivial solutions during self-supervised learning. Then Section \ref{sec:StackOfDif} evaluates the effectiveness of our method in encoding the spatiotemporal context of a scene from single input frames. Section \ref{sec:CohenenceEval} explores the effects of using a higher order loss function both in terms of forming a temporally coherent embedding space and for downstream action recognition. In Section \ref{sec:Kinetics} we demonstrate the advantages and challenges of pre-training with larger and more diverse datasets. Then in Section \ref{sec:SOA} we compare the downstream performance of our method with competing state-of-the-art methods. Finally in Section \ref{sec:Visual} we present low dimensional mappings of several learned embedding spaces in order to visually inspect the structure given by our temporal coherency objective.

\subsection{Impact of Auxiliary Task}
\label{sec:Auxiliary}
\begin{table}
    \caption{A comparison of the performance of our approach with and without the rotation auxiliary loss.}
    \centering 
    \begin{tabular}{llp{3cm}|c}
        \hline \hline 
        \textbf{Network} & \textbf{Train Set} & \textbf{Pre-training} &\textbf{UCF101}\\
        \hline \hline 
        2D R-18 & UCF101 & Random Initialisation & 41.5 \\
        \hline
        2D R-18 & UCF101 & Temporal Coherency Loss & 62.3  \\
        \hline 
        2D R-18 & UCF101 & Temporal Coherency Loss + Rotation Loss & 68.2\\
        \hline 
    \end{tabular}
    \label{tab:init_methods}
\end{table}

Table \ref{tab:init_methods} details the performance achieved by fine-tuned networks on an action recognition task when starting from different baselines. Our self-supervised pre-training shows a dramatic improvement of 21\% over random initialization. Furthermore, the addition of the auxiliary task to the self-supervised pre-training provides further improvement of 5.9\% to the performance on the downstream task. This demonstrates the susceptibility for our self-supervised learning method to learn trivial solutions such as optical flow during the pre-training, reducing the transferability of the leaned embeddings to other downstream tasks. By adding the rotation sub-task, we restrict the embedding space from being able to learn these trivial solutions and see an additional boost to downstream performance.

\begin{table}
    \caption{A comparison of TCE performance when trained with RGB streams or stack-of-differences inputs.}
    \centering
    \begin{tabular}{lll|c}
        \hline \hline 
        \textbf{Network} & \textbf{Train Set} & \textbf{Input} & \textbf{UCF101} \\
        \hline \hline 
        2D R-18 & UCF101 & RGB & 68.2 \\
        \hline 
        2D R-18 & UCF101 & Stack-of-Differences & 70.5 \\
        \hline 

    \end{tabular}
    \label{tab:inputs}
\end{table}

\subsection{Network Input}
\label{sec:StackOfDif}

Several previous works \cite{fernando2017self,kim2019self, lee2017unsupervised} have leveraged the advantages of using a stack-of-differences input in order to capture the temporal dynamics of the video during fine-tuning and classification. In Table \ref{tab:inputs} we compare our baseline method to one using the stack-of-differences approach for the downstream action recognition task. For our method, stack-of-differences yields a 2.3\% performance improvement. This is substantially smaller than the roughly 10\% improvement seen for other methods \cite{kim2019self,lee2017unsupervised}. This demonstrates that our coherency objective has already encoded these temporal dynamics into the embedding space, and so the application of stack-of-differences inputs has a reduced benefit for action recognition.

\subsection{Temporal Coherency and Higher Order Objectives}
\label{sec:CohenenceEval}

An appropriate measure is required for us to quantitatively evaluate the effectiveness of our methodology at forming temporally coherent embeddings of video frame sequences. As such we apply two methods for measuring how a discretely sampled curve deviates from a linear trajectory.
For a video $\mathcal{V}$ containing a sequence of $T$ frame embeddings $\{\mathbf{x}_1,...,\mathbf{x}_T\} \in X$ the Total Absolute Curvature (TAC) is given by
\begin{align}
    TAC = \sum_{i=2}^{T-1} \left|\arccos{\frac{(\mathbf{x}_i - \mathbf{x}_{i-1}).(\mathbf{x}_{i+1} - \mathbf{x}_i)}{||\mathbf{x}_i - \mathbf{x}_{i-1}||\,||\mathbf{x}_{i+1} - \mathbf{x}_i||}}\right|
    \label{eq:TAC}
\end{align}
and we also define the Maximum Absolute Curvature (MAC) to be
\begin{align}
    MAC = \max_{i=2,...,T-1} \left|\arccos{\frac{(\mathbf{x}_i - \mathbf{x}_{i-1}).(\mathbf{x}_{i+1} - \mathbf{x}_i)}{||\mathbf{x}_i - \mathbf{x}_{i-1}||\,||\mathbf{x}_{i+1} - \mathbf{x}_i||}}\right|.
    \label{eq:MAC}
\end{align}

Here we also explore extending our coherency objective to second-order, so that the optimization goal is changed from clustering temporally adjacent frame embeddings to also clustering the differences between those embeddings. Clustering the differences between embeddings should ensure that the trajectory of video embeddings does not significantly vary over short time periods. Our second-order temporal coherency objective is given as

\begin{align}
    \mathcal{L}_{2^{\mathrm{nd}}} = - \mathbb{E} \left[ \log{} \frac{ e^{s_2(f_t,f_{t+1}, f_{t+2})} } { e^{s_2(f_t,f_{t+1}, f_{t+2})}+\sum_{\mathcal{N}_2}{e^{s_2(f_t, f_{t+1}, f_{n})}} } \right]
    \label{eq:SecondOrder}
\end{align}

where $s_2(f_{i}, f_{i+1}, f_{i+2}) = s(f_{i+1} - f_{i}, f_{i+2} - f_{i+1})$. Importantly, the negative examples used in calculating the second-order cross-entropy loss are sampled within the same video as the positives. This is because we believe that sampling negative examples from other videos does not provide sufficiently difficult negative examples to learn from.

\begin{figure}
	\centering
	\begin{tabular}{cc}
	\begin{tikzpicture}
    \begin{axis}[width=0.45\linewidth, label style={font=\small}, tick label style={font=\small},
        legend style={nodes={scale=0.75, transform shape}},
        grid=both,major grid style={gray!30},minor grid style={dashed},minor tick num=1,
        xmin=0,ymin=270,xmax=6,ymax=278,
		legend cell align=left,legend style={at={(1,1.6)}},
		xlabel=Pre-training Epoch,ylabel=Total Absolute Curvature]
			\addplot[mark=square*,black,thick,mark options={draw=black,fill=myOrange,scale=1.5},    			error bars/.cd ,y fixed, y dir=both, y explicit]
			table[x=chkpt,y=avgTAC1st,col sep=comma,skip coords between index={5}{10}]{figures/TAC_graph.csv};
			\addplot[mark=triangle*,black,thick,mark options={draw=black,fill=myGreen,scale=1.8},
    			error bars/.cd,y fixed,y dir=both,y explicit]
    			table[x=chkpt,y=avgTACwt,col sep=comma,skip coords between index={5}{10}] {figures/TAC_graph.csv};
		\legend{First Order Only,First + Second Order}
	\end{axis}
	\end{tikzpicture} &
	\begin{tikzpicture}
	\begin{axis}[width=0.45\linewidth, label style={font=\small}, tick label style={font=\small},
        legend style={nodes={scale=0.75, transform shape}},
        grid=both,major grid style={gray!30},minor grid style={dashed},minor tick num=1,
        xmin=0,ymin=2.1,xmax=10,ymax=2.5,
		legend cell align=left,legend style={at={(1,1.6)}},
		xlabel=Pre-training Epoch,ylabel=Maximum Absolute Curvature]
			\addplot[mark=square*,black,thick,mark options={draw=black,fill=myOrange,scale=1.5},
    			error bars/.cd,y fixed,y dir=both,y explicit]
    			table[x=chkpt,y=avgMAC1st,col sep=comma] {figures/MAC_graph.csv};
			\addplot[mark=triangle*,black,thick,mark options={draw=black,fill=myGreen,scale=1.8},
    			error bars/.cd,y fixed,y dir=both,y explicit]
    			table[x=chkpt,y=avgMACwt,col sep=comma] {figures/MAC_graph.csv};
		\legend{First Order Only,First + Second Order}
	\end{axis}
	\end{tikzpicture}
    \end{tabular}
	\caption{A comparison of TAC and MAC scores during self-supervised pre-training with and without a second order loss.}
	\label{fig:tac_graph}
\end{figure}
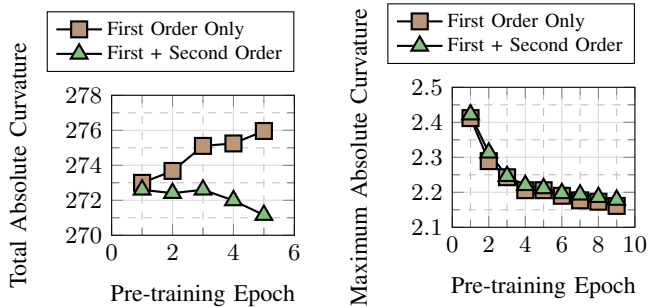

Figure \ref{fig:tac_graph} presents TAC and MAC measures across self-supervised pre-training runs using our proposed loss functions. Each data point is an average of the TACs or MACs computed on 375 videos sampled evenly across all classes of the UCF101 test set. We report results for the first order loss and a combined loss using the first and second order losses in a $5:1$ weight ratio. The TAC plot shows that the addition of a second order loss acts to continuously reduce the curvature of the video trajectories. As such, the combined higher order loss is more effective at improving the temporal coherency of the video embeddings. However, both sets of MAC results decrease throughout the self-supervised pre-training. This indicates that the least temporally coherent frames within each video are still being moved towards a consistent trajectory, even when only the first order loss is applied.

Lastly in Table \ref{tab:ablations} we evaluate these trained embedding spaces on the downstream action recognition task. Here we see that the addition of a higher order loss results in minor decreased performance on the downstream task. From this we conclude that while our proposed loss functions are successfully forming embedding spaces with increasing temporal coherency, this property is not a perfect proxy for the downstream task being evaluated. And as such, it is possible to push the temporal coherency constraints too far in the context of action recognition.

\begin{table}[h]
    \centering
    \caption{A comparison of TCE performance when trained with and without enforcing a second order loss.}
    \begin{tabular}{lc}
        \hline \hline 
         \textbf{Method} \hspace{10mm} & \hspace{5mm}\textbf{UCF101}\hspace{5mm} \\
         \hline \hline 
         First Order Only &  68.2\\
         \hline
         First + Second Order & 66.88\\
         \hline 
    \end{tabular}
    \label{tab:ablations}
\end{table}

\subsection{Benefits of Large Datasets}
\label{sec:Kinetics}
\begin{table}[h]
    \caption{A comparison of TCE performance when pre-trained on a larger dataset with semi-hard mining.}
    \centering
    \begin{tabular}{llc|c}
        \hline \hline 
        \textbf{Network} & \textbf{Train Set} & \textbf{Semi-Hard Mining} & \textbf{UCF101} \\
        \hline \hline 
        2D R-18 & UCF101 & \xmark & 68.2 \\
        \hline 
        2D R-18 & Kinetics400 & \xmark & 65.34 \\
        \hline 
        2D R-18 & Kinetics400 & \cmark & 68.75 \\
        \hline 
        2D R-50 & Kinetics400 & \cmark & 72.14 \\
        \hline 

    \end{tabular}
    \label{tab:scaling}
\end{table}

In Table \ref{tab:scaling} we demonstrates the scalability of TCE to larger datasets with results on Kinetics400. However a naive transition to the large dataset results in reduced performance on the downstream action recognition task. Then with the addition of our semi-hard mining method from Section \ref{sec:mining}, we are instead able to achieve an improvement in performance. This result emphasises the importance of adaptive negative sampling when scaling to larger datasets, and suggests that the vanishing gradient problem is relevant to our temporal coherence loss. Lastly, we achieve a significant performance improvement when we move from pre-training on 2D ResNet-18 to 2D ResNet-50. This result reconfirms the findings of \cite{hara2018can} in that higher-capacity networks are more suited for learning from large datasets. However, it also demonstrates similar performance is achievable based on 2D-CNN SSL solutions on spatiotemporal datasets without relying on the additional representation capacity of 3D convolutions. This gives our approach several inherent advantages in terms of pre-training time, computational resources required and inference speed.

\subsection{Comparison to State-of-the-Art}
    \label{sec:SOA}

      
\begin{table*}[t]
    \centering
        \caption{Top-1 accuracy for action recognition on UCF101 and HMDB51 datasets. Rows are ordered by UCF101 performance.  For fair comparison, we exclude methods which use additional modalities such as optical flow or audio as network inputs. $^{+}$Results reported on train/test split 1 of UCF101. $^{\dagger}$Modified Network Architecture}
        \begin{tabular}{lllclcl}
        \hline \hline 
         Method & Backbone & Params & 2D-CNN & Pre-Training & UCF101(\%) & HMDB51(\%)\\
         \hline \hline 
         3DRotNet \cite{jing2018selfsupervised} & 3D ResNet-18 & $34\times10^6$& \xmark & Kinetics400 & 62.9~~  & 33.7 \\
         3DCubicPuzzles \cite{kim2019self} & 3D ResNet-18 & $34\times10^6$& \xmark & Kinetics400 & 65.8~~ & 33.7 \\
         DPC \cite{han2019video} & 3D ResNet-18$^{\dagger}$ & $14\times10^6$& \xmark & Kinetics400 & 68.2~~ & 34.5 \\
         \textbf{TCE (Ours)} & 2D ResNet-18 & $11\times10^6$ & \cmark & Kinetics400 & 68.8$^+$ & 34.2 \\ 
         \textbf{TCE (Ours)} & 2D ResNet-50 & $23\times10^6$ & \cmark & Kinetics400 & 71.2~~ & \textbf{36.6} \\ 
         DPC \cite{han2019video} & 3D ResNet-34$^{\dagger}$ & $33\times10^6$& \xmark & Kinetics400 & \textbf{75.7}~~ & 35.7 \\
         \hline
         Motion \& Appearance \cite{wang2019selfsupervised}& C3D &$11\times10^6$ & \xmark & UCF101 & 48.6~~  & 20.3 \\
         Shuffle and Learn \cite{misra2016shuffle} & AlexNet & $61\times10^6$& \cmark & UCF101 & 50.9$^+$ & 19.8 \\
         VideoGAN \cite{vondrick2016generating} & C3D & $11\times10^6$& \xmark & UCF101 & 52.1~~ & - \\
         Arrow of time \cite{wei2018learning}& AlexNet & $61\times10^6$& \cmark & UCF101 & 55.3~~ & - \\
         CMC \cite{tian2019contrastive} & CaffeNet $\times 2^{*}$ &$58\times10^6\times 2$ & \cmark & UCF101 & 55.3~~ & - \\
         OPN \cite{lee2017unsupervised} &  VGG-M-2048 & $8.6\times10^6$& \cmark & UCF101 & 59.8~~ & 23.8\\

         DPC \cite{han2019video} & 3D ResNet-18$^{\dagger}$ & $14\times10^6$& \xmark & UCF101 & 60.6$^+$ & - \\
         Skip-Clip \cite{el2019skip} & 3D ResNet-18 & $34\times10^6$& \xmark & UCF101  & 64.4$^+$ & - \\ 
         Video Clip Ordering \cite{xu2019self} & R3D & $14 \times 10^{6}$ & \xmark & UCF101 & 64.9$^+$ & 29.5 \\
         \textbf{\textit{TCE} (Ours)} & 2D ResNet-18 & $11\times10^6$ & \cmark & UCF101 & \textbf{68.2}$^+$ & \textbf{31.7}\\
         \hline 
        \end{tabular}
    \label{tab:soa_table}
\end{table*}

In Table \ref{tab:soa_table} we compare the results of our pre-training method against other state-of-the-art results. Unless noted otherwise the results in this table are averaged over all three train-test splits of the UCF101 dataset, unlike our previous experiments which are performed only on the first split.  When considering only self-supervised pre-training on UCF101, \textit{TCE} outperforms all other methods including several that utilise the additional capacity and have the additional inference times associated with 3D-CNN backbones. Then in the context of self-supervised pre-training on Kinetics400, \textit{TCE} outperforms all other methods except for DPC \cite{han2019video} when evaluating on UCF101 and our method sets a new state-of-the-art performance for HMDB51. It is worth noting that the results from DPC are generated using a much higher capacity network and took six weeks to train, while our method is implemented on a smaller capacity network and was trained for only 72 hours. In addition to increased pre-training time, using larger networks also typically result in longer inference times. The same comparisons can be made to a lesser degree between our Kinetics400 results using different backbones. Based on our strong results, we argue that contrary to the current direction of the field, 2D-CNNs are still capable of learning sufficiently complex spatiotemporal embeddings for many downstream tasks.
\subsection{Visual Inspection of Temporal Coherency}
\label{sec:Visual}
In this section we qualitatively evaluate the effectiveness of \textit{TCE} in creating a temporally coherent embedding space. We visualise our embedding spaces by using t-SNE to learn a non-linear dimension reduction of frame embeddings from a single video. This video was selected from the Bowling class of the UCF101 validation set due to the visually distinct actions of back swinging, releasing the ball, following the ball and the pins falling. Once reduced to two dimensions, we plot the resulting values such that each point represents a single video frame. We apply a color map to show the temporal order of frames within the embedded video.

Figure \ref{fig:training_prog}, seen at the top of this paper, visualises the evolution of our embedding space over the course of unsupervised training. This figure reinforces that our training is continuously increasing the temporal coherency of the embedding space across epochs; from negligible coherency for random weights, to partial coherency at 25 epochs and then very strong coherency at 50 epochs. This result supports our quantitative analysis in Section \ref{sec:CohenenceEval} that demonstrated that we are continuously increasing the temporal coherency through a decreasing maximum absolute curvature.


\begin{figure*}[h]
  \subfloat[][\textit{TCE} (Ours)] {
    \centering
    \includegraphics[width=0.28\textwidth]{figures/trainingProgression/kinetics_ckpt_50.png}}
  \subfloat[][ImageNet Pre-Training] {
    \centering
    \includegraphics[width=0.28\textwidth]{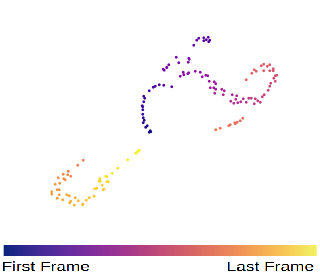}}
    \subfloat[][DPC] {
    \centering
    \includegraphics[width=0.28\textwidth]{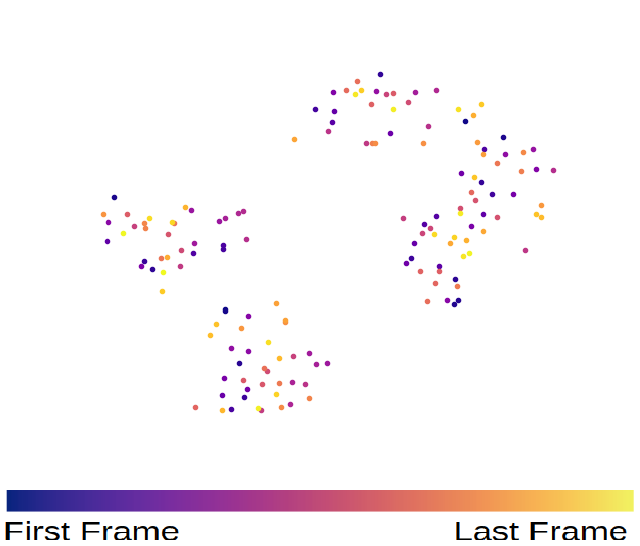}}
  \caption{A comparison between frame embeddings generated with (a) \textit{TCE}; (b) ImageNet Pre-Training; and (c) DPC~\cite{han2019video}.}
  \label{fig:compare_embed}
\end{figure*}

In Figure \ref{fig:compare_embed} we compare visualisations of our embedding space against an ImageNet pre-trained checkpoint and the DPC context representation \cite{han2019video}. For DPC, five consecutive video blocks were sampled and then extracted as feature maps. These feature maps were then aggregated into context representation and finally pooled into vectors. From visual inspection of the three methods, we see that our method is the only one to form a coherent path with no major discontinuities. This suggests that our method is generating embeddings with greater temporal coherency than both ImageNet and DPC.

\label{sec: Conclusion}
\section{Conclusion}

In this paper, we have presented \textit{TCE: Temporally Coherent Embeddings}, an effective 2D-CNN approach for self-supervised representation learning of spatiotemporal features from videos. We train our model in a self-supervised manner by leveraging the temporal information embedded in video data and enforcing coherency in the embedding space. Our first order temporal coherency objective encourages the relative attraction and separation of an anchor frame, an adjacent frame and N negative frames sampled from other videos. Additionally, we explored measures for evaluating the effectiveness of our coherency objective and its extension to a second order objective that can further increase the temporal coherence of our learned embeddings.

We evaluated our learned representations on a downstream action recognition task using multiple challenging benchmarks (Kinetics400, UCF101, HMDB51). With these we have empirically demonstrated the importance of auxiliary tasks for avoiding the learning of trivial solutions and the importance of semi-hard mining for effective negative selection when learning from larger datasets using our temporal coherency objective. In combination these components resulted in our method outperforming all previous state-of-the-art 2D-CNN and 3D-CNN self-supervised learning for action recognition methods that have been pre-trained on UCF101, by at least $3.3\%$ with UCF101 and $2.2\%$ with HMDB51. For methods pre-trained on Kinetics400 we present the only competitive approach using a 2D-CNN.  Finally, our method outperforms all but one of the competing methods on on UCF101 and sets a new state-of-the-art performance for HMDB51. This gives our method the added advantages of shorter training and inference times with less computational hardware, while also demonstrating that 2D-CNNs can still effectively encode spatiotemporal information.

Overall, our analysis has demonstrated that our methodology can deliver competitive generalization results without the complexity of 3D-CNN network architecture. As such, we conclude that explicitly enforcing temporal coherency between nearby frame embeddings is a powerful learning self-supervised training signal. In applying our learned embeddings to the downstream action task, we are able to leverage smaller computational training costs and are faster at inference time. Interestingly, we have also found that there could be such a thing as too much temporal coherency in the context of this particular downstream task. As such, in future work we plan to investigate the application of pre-trained \textit{TCE} for downstream tasks that require a higher level of temporal understanding, such as action anticipation or phase classification.

\balance

\bibliographystyle{IEEEtran}
\bibliography{IEEEabrv, bib}

\begin{thebibliography}{10}
\providecommand{\url}[1]{#1}
\csname url@samestyle\endcsname
\providecommand{\newblock}{\relax}
\providecommand{\bibinfo}[2]{#2}
\providecommand{\BIBentrySTDinterwordspacing}{\spaceskip=0pt\relax}
\providecommand{\BIBentryALTinterwordstretchfactor}{4}
\providecommand{\BIBentryALTinterwordspacing}{\spaceskip=\fontdimen2\font plus
\BIBentryALTinterwordstretchfactor\fontdimen3\font minus
  \fontdimen4\font\relax}
\providecommand{\BIBforeignlanguage}[2]{{%
\expandafter\ifx\csname l@#1\endcsname\relax
\typeout{** WARNING: IEEEtran.bst: No hyphenation pattern has been}%
\typeout{** loaded for the language `#1'. Using the pattern for}%
\typeout{** the default language instead.}%
\else
\language=\csname l@#1\endcsname
\fi
#2}}
\providecommand{\BIBdecl}{\relax}
\BIBdecl

\bibitem{lotter2016deep}
W.~Lotter, G.~Kreiman, and D.~Cox, ``Deep predictive coding networks for video
  prediction and unsupervised learning,'' in \emph{International Conference on
  Learning Representations}, 2017.

\bibitem{mathieu2015deep}
M.~Mathieu, C.~Couprie, and Y.~LeCun, ``Deep multi-scale video prediction
  beyond mean square error,'' \emph{ICLR}, 2016.

\bibitem{srivastava2015unsupervised}
N.~Srivastava, E.~Mansimov, and R.~Salakhudinov, ``Unsupervised learning of
  video representations using lstms,'' in \emph{International conference on
  machine learning}, 2015, pp. 843--852.

\bibitem{vondrick2016generating}
C.~Vondrick, H.~Pirsiavash, and A.~Torralba, ``Generating videos with scene
  dynamics,'' in \emph{Advances in neural information processing systems},
  2016, pp. 613--621.

\bibitem{han2019video}
T.~Han, W.~Xie, and A.~Zisserman, ``Video representation learning by dense
  predictive coding,'' in \emph{Proceedings of the IEEE International
  Conference on Computer Vision Workshops}, 2019, pp. 0--0.

\bibitem{vondrick2016anticipating}
C.~Vondrick, H.~Pirsiavash, and A.~Torralba, ``Anticipating visual
  representations from unlabeled video,'' in \emph{Proceedings of the IEEE
  Conference on Computer Vision and Pattern Recognition}, 2016, pp. 98--106.

\bibitem{tran2018closer}
D.~Tran, H.~Wang, L.~Torresani, J.~Ray, Y.~LeCun, and M.~Paluri, ``A closer
  look at spatiotemporal convolutions for action recognition,'' in
  \emph{Proceedings of the IEEE conference on Computer Vision and Pattern
  Recognition}, 2018, pp. 6450--6459.

\bibitem{el2019skip}
A.~El-Nouby, S.~Zhai, G.~W. Taylor, and J.~M. Susskind, ``Skip-clip:
  Self-supervised spatiotemporal representation learning by future clip order
  ranking,'' \emph{arXiv preprint arXiv:1910.12770}, 2019.

\bibitem{fernando2017self}
B.~Fernando, H.~Bilen, E.~Gavves, and S.~Gould, ``Self-supervised video
  representation learning with odd-one-out networks,'' in \emph{Proceedings of
  the IEEE conference on computer vision and pattern recognition}, 2017, pp.
  3636--3645.

\bibitem{kim2019self}
D.~Kim, D.~Cho, and I.~S. Kweon, ``Self-supervised video representation
  learning with space-time cubic puzzles,'' in \emph{Proceedings of the AAAI
  Conference on Artificial Intelligence}, vol.~33, 2019, pp. 8545--8552.

\bibitem{lee2017unsupervised}
H.-Y. Lee, J.-B. Huang, M.~Singh, and M.-H. Yang, ``Unsupervised representation
  learning by sorting sequences,'' in \emph{Proceedings of the IEEE
  International Conference on Computer Vision}, 2017, pp. 667--676.

\bibitem{misra2016shuffle}
I.~Misra, C.~L. Zitnick, and M.~Hebert, ``Shuffle and learn: unsupervised
  learning using temporal order verification,'' in \emph{European Conference on
  Computer Vision}.\hskip 1em plus 0.5em minus 0.4em\relax Springer, 2016, pp.
  527--544.

\bibitem{xu2019self}
D.~Xu, J.~Xiao, Z.~Zhao, J.~Shao, D.~Xie, and Y.~Zhuang, ``Self-supervised
  spatiotemporal learning via video clip order prediction,'' in
  \emph{Proceedings of the IEEE Conference on Computer Vision and Pattern
  Recognition}, 2019, pp. 10\,334--10\,343.

\bibitem{bachman2019learning}
P.~Bachman, R.~D. Hjelm, and W.~Buchwalter, ``Learning representations by
  maximizing mutual information across views,'' in \emph{Advances in Neural
  Information Processing Systems}, 2019, pp. 15\,535--15\,545.

\bibitem{chen2020simple}
T.~Chen, S.~Kornblith, M.~Norouzi, and G.~Hinton, ``A simple framework for
  contrastive learning of visual representations,'' \emph{arXiv preprint
  arXiv:2002.05709}, 2020.

\bibitem{sermanet2018time}
P.~Sermanet, C.~Lynch, Y.~Chebotar, J.~Hsu, E.~Jang, S.~Schaal, S.~Levine, and
  G.~Brain, ``Time-contrastive networks: Self-supervised learning from video,''
  in \emph{2018 IEEE International Conference on Robotics and Automation
  (ICRA)}.\hskip 1em plus 0.5em minus 0.4em\relax IEEE, 2018, pp. 1134--1141.

\bibitem{tian2019contrastive}
Y.~Tian, D.~Krishnan, and P.~Isola, ``Contrastive multiview coding,''
  \emph{arXiv preprint arXiv:1906.05849}, 2019.

\bibitem{zhang2017split}
R.~Zhang, P.~Isola, and A.~A. Efros, ``Split-brain autoencoders: Unsupervised
  learning by cross-channel prediction,'' in \emph{Proceedings of the IEEE
  Conference on Computer Vision and Pattern Recognition}, 2017, pp. 1058--1067.

\bibitem{gan2018geometry}
C.~Gan, B.~Gong, K.~Liu, H.~Su, and L.~J. Guibas, ``Geometry guided
  convolutional neural networks for self-supervised video representation
  learning,'' in \emph{Proceedings of the IEEE Conference on Computer Vision
  and Pattern Recognition}, 2018, pp. 5589--5597.

\bibitem{jayaraman2016slow}
D.~Jayaraman and K.~Grauman, ``Slow and steady feature analysis: higher order
  temporal coherence in video,'' in \emph{Proceedings of the IEEE Conference on
  Computer Vision and Pattern Recognition}, 2016, pp. 3852--3861.

\bibitem{tschannen2019mutual}
M.~Tschannen, J.~Djolonga, P.~K. Rubenstein, S.~Gelly, and M.~Lucic, ``On
  mutual information maximization for representation learning,'' in
  \emph{International Conference on Learning Representations}, 2020.

\bibitem{milbich2017unsupervised}
T.~Milbich, M.~Bautista, E.~Sutter, and B.~Ommer, ``Unsupervised video
  understanding by reconciliation of posture similarities,'' in
  \emph{Proceedings of the IEEE International Conference on Computer Vision},
  2017, pp. 4394--4404.

\bibitem{wang2015unsupervised}
X.~Wang and A.~Gupta, ``Unsupervised learning of visual representations using
  videos,'' in \emph{Proceedings of the IEEE International Conference on
  Computer Vision}, 2015, pp. 2794--2802.

\bibitem{mnih2013learning}
A.~Mnih and K.~Kavukcuoglu, ``Learning word embeddings efficiently with
  noise-contrastive estimation,'' in \emph{Advances in neural information
  processing systems}, 2013, pp. 2265--2273.

\bibitem{gutmann2010noise}
M.~Gutmann and A.~Hyv{\"a}rinen, ``Noise-contrastive estimation: A new
  estimation principle for unnormalized statistical models,'' in
  \emph{Proceedings of the Thirteenth International Conference on Artificial
  Intelligence and Statistics}, 2010, pp. 297--304.

\bibitem{schroff2015facenet}
F.~Schroff, D.~Kalenichenko, and J.~Philbin, ``Facenet: A unified embedding for
  face recognition and clustering,'' in \emph{Proceedings of the IEEE
  conference on computer vision and pattern recognition}, 2015, pp. 815--823.

\bibitem{harwood2017smart}
B.~Harwood, V.~Kumar~BG, G.~Carneiro, I.~Reid, and T.~Drummond, ``Smart mining
  for deep metric learning,'' in \emph{Proceedings of the IEEE International
  Conference on Computer Vision}, 2017, pp. 2821--2829.

\bibitem{soomro2012UCF101}
K.~Soomro, A.~R. Zamir, and M.~Shah, ``{UCF101}: A dataset of 101 human actions
  classes from videos in the wild,'' \emph{arXiv preprint arXiv:1212.0402},
  2012.

\bibitem{kuehne2011hmdb}
H.~Kuehne, H.~Jhuang, E.~Garrote, T.~Poggio, and T.~Serre, ``{HMDB}: a large
  video database for human motion recognition,'' in \emph{2011 International
  Conference on Computer Vision}.\hskip 1em plus 0.5em minus 0.4em\relax IEEE,
  2011, pp. 2556--2563.

\bibitem{kay2017kinetics}
W.~Kay, J.~Carreira, K.~Simonyan, B.~Zhang, C.~Hillier, S.~Vijayanarasimhan,
  F.~Viola, T.~Green, T.~Back, P.~Natsev, M.~Suleyman, and A.~Zisserman, ``The
  kinetics human action video dataset,'' 2017.

\bibitem{he2016deep}
K.~He, X.~Zhang, S.~Ren, and J.~Sun, ``Deep residual learning for image
  recognition,'' in \emph{Proceedings of the IEEE conference on computer vision
  and pattern recognition}, 2016, pp. 770--778.

\bibitem{gidaris2018unsupervised}
S.~Gidaris, P.~Singh, and N.~Komodakis, ``Unsupervised representation learning
  by predicting image rotations,'' in \emph{International Conference on
  Learning Representations}, 2018.

\bibitem{hara2018can}
K.~Hara, H.~Kataoka, and Y.~Satoh, ``Can spatiotemporal 3{D} {CNNs} retrace the
  history of 2{D} cnns and imagenet?'' in \emph{Proceedings of the IEEE
  conference on Computer Vision and Pattern Recognition}, 2018, pp. 6546--6555.

\bibitem{jing2018selfsupervised}
L.~Jing, X.~Yang, J.~Liu, and Y.~Tian, ``Self-supervised spatiotemporal feature
  learning via video rotation prediction,'' \emph{arXiv preprint
  arXiv:1811.11387}, 2018.

\bibitem{wang2019selfsupervised}
J.~Wang, J.~Jiao, L.~Bao, S.~He, Y.~Liu, and W.~Liu, ``Self-supervised
  spatio-temporal representation learning for videos by predicting motion and
  appearance statistics,'' in \emph{Proceedings of the IEEE Conference on
  Computer Vision and Pattern Recognition}, 2019, pp. 4006--4015.

\bibitem{wei2018learning}
D.~Wei, J.~J. Lim, A.~Zisserman, and W.~T. Freeman, ``Learning and using the
  arrow of time,'' in \emph{Proceedings of the IEEE Conference on Computer
  Vision and Pattern Recognition}, 2018, pp. 8052--8060.

\end{thebibliography}

\end{document}